\documentclass[conference]{IEEEtran}
\IEEEoverridecommandlockouts
\usepackage{cite}
\usepackage{amsmath,amssymb,amsfonts}
\usepackage{algorithmic}
\usepackage{graphicx}
\usepackage{textcomp}
\usepackage{multirow}
\usepackage{xcolor}
\def\BibTeX{{\rm B\kern-.05em{\sc i\kern-.025em b}\kern-.08em
    T\kern-.1667em\lower.7ex\hbox{E}\kern-.125emX}}

\makeatletter
\g@addto@macro\normalsize{%
  \setlength{\abovedisplayskip}{-3pt}%
  \setlength{\belowdisplayskip}{0pt}%
  \setlength{\abovedisplayshortskip}{-3pt}%
  \setlength{\belowdisplayshortskip}{0pt}%
}
\makeatother

\begin{document}

\title{Beyond Line of Sight: Hybrid Validation of V2X Collective Perception in Complex Scenarios\\
\thanks{This research has been conducted as part of the EVENTS project, which is funded by the European Union, under grant agreement No 101069614. Views and opinions expressed are however those of the author(s) only and do not necessarily reflect those of the European Union or European Commission. Neither the European Union nor the granting authority can be held responsible for them.}
}

\author{
\IEEEauthorblockN{
Markos Antonopoulos, Anastasia Bolovinou, Bill Roungas, Elena Daskalaki, Angelos Amditis}
\IEEEauthorblockA{\textit{Institute of Communication and Computer Systems}\\
Athens, Greece\\
\{markos.antonopoulos, anastasia.bolovinou, vroungas, e.daskalaki, a.amditis\}@iccs.gr}
}

\maketitle

\begin{abstract}
This paper introduces a probabilistic framework and hybrid validation methodology for V2X-enabled Collective Perception (CP) in complex traffic scenarios. The proposed Bayesian fusion algorithm extends the perceptual horizon of connected and autonomous vehicles by integrating heterogeneous sensor observations from multiple agents into a shared probabilistic occupancy grid. Each cell of this grid encapsulates both occupancy likelihood and uncertainty, enabling explainable and trustworthy situational awareness beyond the ego vehicle's field of view. To bridge the gap between simulation and real-world evaluation, a hybrid testing framework is developed, combining CARLA-based virtual environments with vehicle-in-the-loop experimentation. Experimental results in a roundabout scenario demonstrate a 260\% increase in field-of-view coverage and a rise in occupied-cell recall from 0.82 (ego-only) to 0.94 (six-agent CP) under nominal localization conditions. Overall, the proposed approach provides a reproducible and interpretable foundation for validating CP systems, supporting the safe and certifiable deployment of cooperative autonomous vehicles.
\end{abstract}


\section{Introduction}\label{sec:intro}

Collective Perception (CP), currently standardized by ETSI as a second generation V2X communication service, is especially promising for Autonomous Vehicles (AVs), as it allows connected AV agents ``see through the eyes of others'' who may share processed sensor data via V2X communication. Its benefit is typically assessed in terms of the increased object update rate, extended field-of-view awareness, and redundancy. In this work, a safety validation proof-of-concept (PoC) for CP scenario-based testing for a roundabout navigation scenario is designed. The PoC includes both connected AVs and non-connected vehicles, assuming connected virtual agents that can exchange ETSI-alike Collective Perception Message (CPM) information. Hence, the objective of this work is to (1) develop an algorithm for fusion of object information coming from multiple observers based on probabilistic scene state estimation via occupancy grid maps, (2) integrate data reliability metrics enabling penalization of CP unreliable data and object associations’ conflict resolution, and (3) explore virtual and hybrid test environments for CP system performance evaluation focusing on augmented perception quality. It should be noted that for reasons of simplicity, no networking aspects are taken into account; ETSI-alike CP messages are assumed available with frequency/delay that can vary.

\section{Literature Review}\label{sec:lit}

Early CP research primarily focused on communication efficiency, including latency and congestion control, to maintain Quality of Service (QoS). However, perception-centered research has reframed CP as a multi-sensor data fusion problem. Bayesian and probabilistic models now dominate the field due to their inherent explainability and ability to quantify uncertainty. Foundational works by \cite{godoy2021grid} and \cite{nuss2018random} introduced grid-based and random finite set (RFS) formulations that support consistent multi-observer fusion, establishing the basis for modern probabilistic CP systems.

Bayesian approaches underpin most current CP frameworks by modeling the environment as a probabilistic occupancy grid. Each grid cell represents the likelihood of occupancy, updated recursively as new observations are received from connected agents. The \textit{Bayesian Occupancy Filter (BOF)} \cite{coue2006bayesian} and its hybrid variants \cite{negre2014hybrid} introduced motion-aware recursive estimation, forming the theoretical basis for real-time perception fusion. These models offer interpretability and quantitative uncertainty metrics—critical for safety validation and fault detection in autonomous systems. Recent developments extend these principles to V2X-enabled cooperative perception. \cite{shan2022probabilistic} proposed a probabilistic V2X fusion framework using Bayesian filtering to integrate uncertain sensor data across vehicles. Similarly, \cite{huang2025v2x} introduced adaptive Bayesian reweighting for multi-agent perception, dynamically adjusting the confidence of individual observations under varying communication conditions. These contributions demonstrate the scalability and robustness of Bayesian schemes for dynamic traffic scenarios, including intersections and roundabouts.

The rapid growth of collaborative perception datasets such as \textit{V2X-Sim} and \textit{OPV2V} \cite{li2022v2x, liu2023survey} has enabled benchmarking of perception fusion strategies across early, intermediate, and late fusion paradigms. These datasets facilitate standardized evaluation of object detection, spatial coverage, and latency-aware performance. To bridge the gap between simulation and field testing, hybrid frameworks have emerged that integrate real and virtual agents through platforms like CARLA, ROS2, and MathWorks RoadRunner. The hybrid approach proposed by \cite{antonopoulos2024roundabout} combines probabilistic Bayesian fusion with a virtual safety validation environment, enabling controlled scenario-based experiments in complex environments such as roundabouts. Similarly, \cite{bolovinou2024virtual} proposed a virtual safety validation architecture for urban CP scenarios, introducing new metrics to assess awareness and reliability in mixed-reality simulations. These frameworks align with the growing emphasis on explainability and repeatability in CP validation.

In \cite{wei2025cooperative, bejarbaneh2024shared, huang2023recent}, 4 converging research directions are identified:

\begin{itemize}
    \item \textbf{Shift toward perception-centric design:} CP research increasingly integrates perception fusion and situational awareness rather than focusing solely on connectivity.
    \item \textbf{Explainable probabilistic reasoning:} Bayesian models provide transparent confidence metrics essential for perception reliability and decision assurance.
    \item \textbf{Hybrid testing environments:} Mixed-reality validation offers safe, reproducible conditions for testing CP under varying noise and occlusion.
    \item \textbf{Standardized datasets and metrics:} Datasets such as V2X-Sim and OPV2V support reproducible benchmarking of fusion performance.
\end{itemize}

Despite these advances, several gaps persist. Most studies isolate communication or perception aspects, and few integrate uncertainty quantification directly into safety validation. Furthermore, hybrid testing infrastructures are not yet standardized, limiting comparability across implementations. The Bayesian CP algorithm and hybrid validation plan proposed in this paper aim to address these challenges by combining interpretable probabilistic fusion with scenario-based evaluation under real and simulated conditions.

\section{Methodology}\label{sec:methodology}

\begin{figure}[htbp]%
\centering
\includegraphics[width=0.4\textwidth]{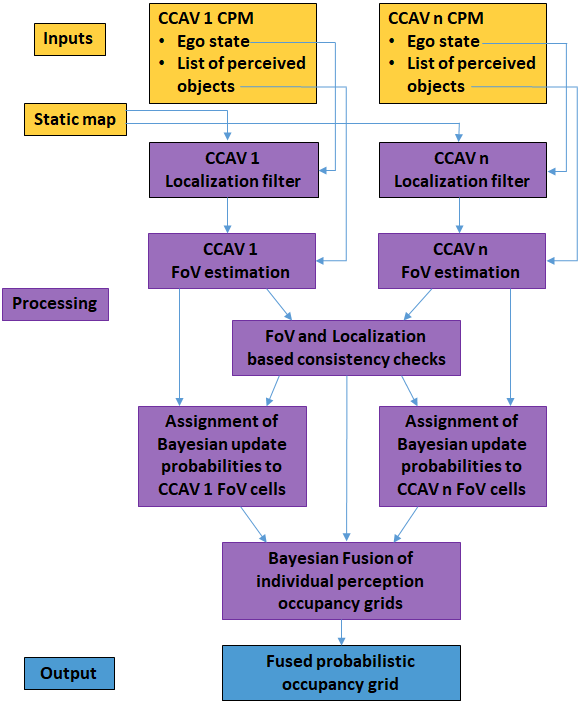}
\caption{The architecture of the CP module.}\label{fig:arch}
\vspace{-5mm}
\end{figure}

This section provides an overview of the core algorithmic steps comprising the proposed CP framework, the architecture of which is depicted in Figure \ref{fig:arch}..

\subsection{Module's Inputs/Outputs}\label{subsec:inputoutput}

The core algorithm of the framework utilizes information originating from CCAVs present in the area under consideration. Specifically, each CCAV is expected to disseminate in real time the following information: \textbf{1. CCAV FoV angle:} Each CCAV’s sensor suite implies a FoV angle for the particular CCAV. For example, a single front camera implies that the CCAV’s FoV angle is equal to the FoV of the camera. In the illustrations in Figure \ref{fig:fov}, a $360^o$ FoV angle camera/lidar sensor suite is considered. \textbf{2. Ego state information:} This information consists of the i. Ego position coordinates in $x$,$y$, ii. Ego speed vector $v_x$, $v_y$, iii. Ego heading (yaw angle). \textbf{3. Observed object information:} This information concerns each one of the distinct objects observed by the CCAV and consists of the i. Object's position coordinates in $x$,$y$, ii. Object's speed vector $v_x$, $v_y$, iii. Object's heading (yaw angle).

\subsection{Algorithmic implementation}\label{subsec:implementation}

We consider a two-dimensional bird’s-eye view of the physical region of interest.  The region is discretized by a rectangular grid of dimensions $N \times N$.  Each cell $i$ ($i = 1, \ldots, N^2$) is associated with a binary random variable $A_i \in \{0, 1\}$, where $A_i = 1$ stands for ``cell $i$ is occupied'' and $A_i = 0$ stands for ``cell $i$ is free''. The resulting $N^2$ probabilities $P(A_i = 1), \; i = 1, \ldots, N^2$, each indicating the probability that the corresponding cell $i$ is occupied, constitute a \textit{probabilistic occupancy grid}.

We assume the presence of both connected, who are able to disseminate information concerning their ego state and perceived objects, according to the specification of inputs, and non-connected vehicles. The ground truth of the scene is a non-random occupancy grid, where for each $i$ ($i = 1, \ldots, N^2$), the corresponding probability $P(A_i = 1)$ is either $0$ or $1$. The algorithm aims to estimate the ground truth in terms of a probabilistic occupancy grid by taking into account the observations and measurements of each individual CCAV and their statistical uncertainties/errors.

\begin{figure}[htbp]%
\vspace{-4mm}
\centering
\includegraphics[width=0.2\textwidth]{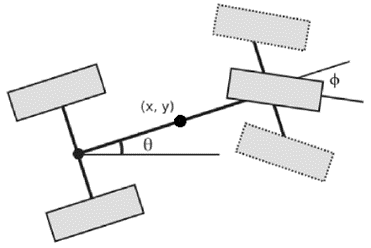}
\caption{Visualization of vehicle dynamics.}\label{fig:dynamics}
\vspace{-2mm}
\end{figure}

\paragraph{Step 1- Particle filter based localization} The first step concerns CCAV localization, i.e., aims to estimate the position of each CCAV. To this end, a properly parametrized particle filter is designed for each CCAV based on the following motion model: $\dot{x} = v \cos\theta - \frac{v \tan\phi}{2} \sin\theta, \dot{y} = v \sin\theta - \frac{v \tan\phi}{2} \cos\theta, \dot{\theta} = \frac{v}{l} \tan\phi$, where $x,y$ are the cartesian coordinates of the center of the CCAV axis, $\theta$ is the heading angle, $\phi$ is the Ackermann steering angle, and $l$ denotes the distance between front and back wheel axes (Figure \ref{fig:dynamics}).

Despite the deterministic nature of the motion model, additive noise terms are introduced to each equation, in order to both make the model more robust and avoid particle degeneration. The measurement model adds gaussian noise to the actual values of $x,y,\theta$ based on the respective measurement error of the devices. Note that for each distinct CCAV, the corresponding filter takes into account only the particular CCAV’s self-reporting measurements of $x,y,\theta$ ignoring measurements originating from the perception of nearby CCAVs. There are two reasons for this choice; first, self-reporting measurements of position, velocity and heading originate from on-board sensors like (possibly differential) GPS and odometry, which are expected to be much less prone to errors and noise. Second, taking into account additional measurements originating from other vehicles would first require clustering the measurements around guessed vehicle positions. However, simulations have shown that due to the unknown number of vehicles and measurement noise, such a clustering step lacks robustness and can lead to unreliable estimations.

\begin{figure}[htbp]%
\centering
\includegraphics[width=0.4\textwidth]{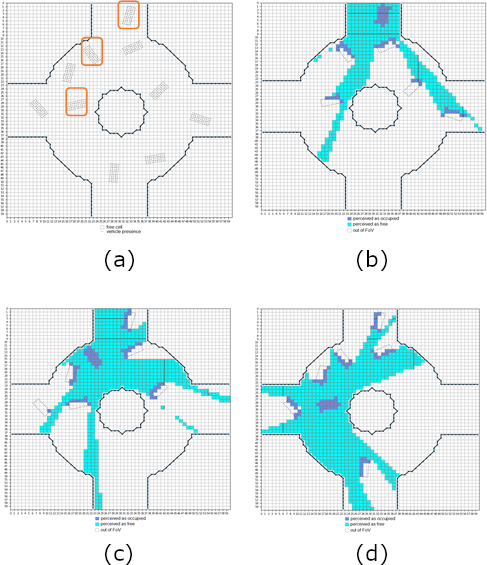}
\caption{FoV estimation for three distinct CCAV agents.}\label{fig:fov}
\vspace{-2mm}
\end{figure}

\paragraph{Step 2 – FoV estimation} The second step estimates the FoV of each CCAV. Each CCAV is placed in its estimated (by step 1) position. Measurements and observations of the CCAV are also placed accordingly. A custom, GPU-implemented algorithm calculates the FoV of the CCAV, i.e., the grid cells visible by the CCAV’s perception plus their perceived occupancy status. Note that prior to the FoV calculation, perceived occupied areas or objects may be dilated, to account for measurement uncertainties. Exemplary outputs of this step are visualized in Figure \ref{fig:fov} (b),(c),(d); ground truth is given in Figure \ref{fig:fov} (a).

\paragraph{Step 3 – Bayesian fusion} The third step fuses the observations of each CCAV aiming to calculate a probabilistic estimation for the occupancy state of each grid cell in the area under consideration. A known individual perception model is assumed for each CCAV, provided in terms of a standard forward sensor model \cite{nuss2018random}, i.e., the four probabilities 
$P(M_i = 0 \mid A_i = 0)$, 
$P(M_i = 1 \mid A_i = 0)$, 
$P(M_i = 0 \mid A_i = 1)$, 
and 
$P(M_i = 1 \mid A_i = 1)$, 
where $A_i \in \{0,1\}$ denotes the random variable ``cell $i$ is actually occupied ($A_i = 1$) or not ($A_i = 0$)''; $M_i \in \{0,1\}$ denotes the random variable ``cell $i$ is perceived as occupied ($M_i = 1$) or not ($M_i = 0$)''.

Taking into account the above probabilities and the output of step 3, Bayes' rule is applied recursively: Let $M_i^k$ be the occupancy state of cell $i$ as perceived by CCAV $k$. For $k = 1$, we have the standard Bayesian update (Equations \ref{eq:bayes_ai1}), where $\alpha = 1$ when cell $i$ is occupied and $\alpha = 0$ when cell $i$ is unoccupied.

\begin{figure*}[!t]
\vspace{2mm}
\normalsize
\begin{equation}
P(A_i = \alpha \mid M_i^1) = 
\frac{P(M_i^1 \mid A_i = 1) P(A_i = 1)}
{P(M_i^1 \mid A_i = 1) P(A_i = 1) + P(M_i^1 \mid A_i = 0) P(A_i = 0)}
\label{eq:bayes_ai1}
\end{equation}
\end{figure*}

These equations calculate $P(A_i = 1 \mid M_i^1)$ and $P(A_i = 0 \mid M_i^1)$, i.e., the posterior occupancy probabilities of cell $i$ conditioned on $M_i^1$. The prior occupancy probabilities $P(A_i = 1)$ and $P(A_i = 0)$ are set initially to $0.5$, reflecting the initial unknown occupancy state of cell $i$. 

Given $k$ independent observations $M_i^1, M_i^2, \ldots, M_i^k$ by $k$ distinct CCAVs on the occupancy state of grid cell $i$, consider the equation \ref{eq:recursion_ai1}, where $\alpha = 1$ when cell $i$ is occupied and $\alpha = 0$ when cell $i$ is unoccupied.

\begin{figure*}[!t]
\normalsize
\begin{equation}
P(A_i = \alpha \mid M_i^1, \ldots, M_i^k) =
\frac{
P(M_i^k \mid A_i = 1) P(A_i = 1 \mid M_i^1, \ldots, M_i^{k-1})
}{
P(M_i^k \mid A_i = 1) P(A_i = 1 \mid M_i^1, \ldots, M_i^{k-1}) + 
P(M_i^k \mid A_i = 0) P(A_i = 0 \mid M_i^1, \ldots, M_i^{k-1})
}
\label{eq:recursion_ai1}
\end{equation}
\end{figure*}

Here, 
$P(A_i = 1 \mid M_i^1, \ldots, M_i^{k-1})$ and 
$P(A_i = 0 \mid M_i^1, \ldots, M_i^{k-1})$ 
are the ``prior'' occupancy probabilities of cell $i$ conditioned on $M_i^1, \ldots, M_i^{k-1}$, 
calculated by the previous recursion steps, 
while 
$P(A_i = 1 \mid M_i^1, \ldots, M_i^k)$ and 
$P(A_i = 0 \mid M_i^1, \ldots, M_i^k)$ 
are the posterior probabilities of cell $i$ conditioned on $M_i^1, \ldots, M_i^k$.  
Note that the result of the entire recursion is invariant to the order in which the measurements 
$M_i^1, M_i^2, \ldots, M_i^k$ are considered.

The recursion is carried out for each grid cell. Each step requires (i) the occupancy probabilities of cell $i$ computed up to step $k-1$, and (ii) the forward sensor model probabilities for agent $k$, to perform the Bayesian update. For cells outside the field of view (FoV) of agent $k$, the probabilities 
$P(M_i = 0 \mid A_i = 0)$, 
$P(M_i = 1 \mid A_i = 0)$, 
$P(M_i = 0 \mid A_i = 1)$, 
and $P(M_i = 1 \mid A_i = 1)$ are set to equal values (any arbitrary number), which simply translates to (again $\alpha = 1$ means that cell $i$ is occupied and $\alpha = 0$ means that cell $i$ is unoccupied):

$P(A_i = 1 \mid M_i^1, \ldots, M_i^k) = P(A_i = 1 \mid M_i^1, \ldots, M_i^{k-1})$

and reflecting the fact that agent k cannot provide information on the occupancy state of the cell, so any prior knowledge remains unchanged.

\begin{figure*}[htbp]%
\centering
\includegraphics[width=0.7\textwidth]{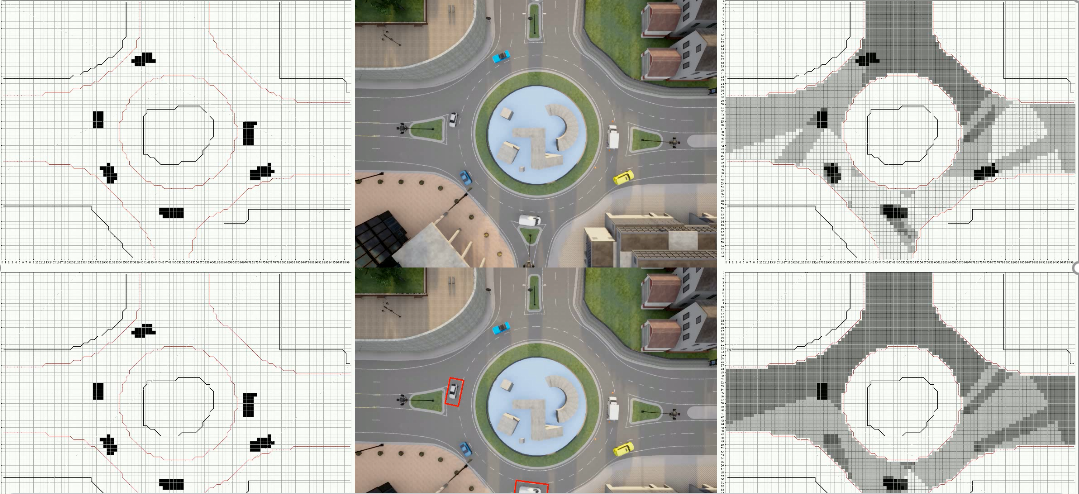}
\caption{Ground truth (left), bird-eye view representation in Carla simulator (middle), and probabilistic occupancy grid (right)}\label{fig:cpmodule}
\vspace{-3mm}
\end{figure*}

The above calculations constitute a probabilistic formalization of the joint consideration of individual observations (if any) regarding the occupancy state of a cell. The output is a probabilistic occupancy grid whose calculation takes into account both each CCAV’s observations plus their corresponding uncertainties, ultimately accounting for the collective perception of the CCAVs, the result of which is depicted in the right illustrations in Figure \ref{fig:cpmodule} (above with 6 connected agents and below with 3 connected agents).
 
\paragraph{Step 4 – Tracking} The process involves a last step envisioned to keep track of the evolution of the occupancy grid through time. Although several methods have been proposed in the literature $[3,5,8,9]$, they are all based in the same heuristic idea; In each step, the occupancy state of each cell is moved to (a) neighboring cell(s) according to their corresponding estimated/measured velocity. The resulting predicted occupancy grid provides the priors for the next time step \cite{coue2006bayesian}. These priors are updated upon arrival of the corresponding measurements as described in step 3. In this work, at each time step, the occupancy state of each cell is considered unknown (i.e., $(A_i=1) = P(A_i=0)=0.5$ ) as described in step 3.

\subsection{Collective perception online reliability assessment}\label{subsec:reliability}

In this section, we argue that the described approach lends itself to straightforward and intuitive derivations of quantitative indicators for assessing  online  the reliability of the output.

\paragraph{CCAV localization} Starting from the CCAVs localization step, a set of reliability indicators for the step’s output can be derived from the estimated covariances of the posterior probabilities of the filter recursions. These covariances characterize the uncertainty ellipse around each estimated CCAV position. In case of Kalman filters, these are the resulting covariance matrices; in \cite{sun2017data} the authors utilize these matrices for conducting pertinent statistical tests. In case of a particle filter, the resulting covariances can be directly calculated from the (resampled) output particle population. 

Thus, estimated covariances indicating uncertainty ellipses above a certain threshold size can be considered unreliable. In principle, one can expect this threshold size to be context dependent. Lower (higher) vehicle congestion at lower (higher) speeds may imply a higher (lower) threshold size. However, and since standard GPS systems provide measurements with a standard deviation of approximately 0.30 m (i.e., the error is within 1m with $>99\%$ probability  and much lower for differential GPS), estimations with  covariances indicating standard deviations above the order of cm can be considered unreliable.

\paragraph{FoV estimation} For each CCAV, this step essentially classifies grid cells into three classes: visible and empty, visible and occupied, and invisible (i.e., cells outside the FoV of the CCAV). This information provides a basis for performing plausibility checks and assessing the consistency of the disseminated individual perception data \cite{zhang2022trust, van2018survey, so2018integrating, kamel2020simulation}. We distinguish two categories of such checks:

\noindent 1) \textbf{Self-contradicting information:} Checks detecting self-contradicting information. In this case, the perception data disseminated by each particular CCAV is checked for internal consistency. This essentially means that the objects claimed to be perceived by the CCAV should be within the reporting CCAV’s FoV. If an object claimed to be perceived by the CCAV is not visible by the estimated location of the CCAV, i.e., if the object is outside the CCAV’s estimated FoV, then one or both of the object’s and CCAV’s location are considered untrustworthy.

\noindent 2) \textbf{All individual perception:} Checks detecting contradicting information across all individual perception data disseminated by the CCAVs. Several checks are possible:

2.1) Each agent claiming to be within the FoV of one or more other agents, should be perceived also by them. In other words, cells claimed to be occupied by an agent that are within the FoV of other agents, should also be perceived as occupied by the other agents.

2.2) Information disseminated from distinct CCAVs concerning regions of intersecting FoVs should not contradict each other. Equivalently, cells perceived by an agent as occupied (or free) that are within the FoV of other agents, should be perceived as such also by the other agents.

In each case, the number of cells with conflicting occupancy information indicates the severity of the respective contradiction. (2.1) and (2.2) can be further categorized to one vs one (i.e., one agent contradicts with one agent) or one vs more (i.e., one agents disagrees with two or more other agents) cases with the latter implying stronger unreliability indications for the non-conforming agent.

\paragraph{Bayesian Fusion} The output of steps 3 and 4 is a probabilistic occupancy grid. Such a concept already divides grid cells in two classes: Cells whose occupancy probability is adequately high or adequately low, and cells whose occupancy probability is neither; The first class indicates cells whose occupancy state is known with high certainty and the second class to cells whose occupancy state is uncertain.

By definition, cells of the second class indicate the locations of the grid for which the occupancy estimates are unreliable. This may be due to missing information (e.g., cells may invisible by every agent) or conflicts between individual perception data of the involved agents as described in the previous section. It is emphasized that in each case, either lack or conflict of available information will be reflected in the bayesian calculation of step 3 and its resulting posterior occupancy probabilities. To illustrate this, consider a case where one CCAV agent fails to detect a present vehicle for the example depicted in Fig 3. Grid cells occupied by this vehicle will appear in the grid with lower occupancy probabilities.

Ultimately, there may be regions (i.e., cells) for which the collective perception output is reliable, and regions for which it is not. Hence, the overall approach does not address the issue of reliability of perception only as a (set of) general metric(s), but considers and accounts for its locality. The locality of (collective) perception reliability is crucial in automated driving applications. While a large number of cells with unreliable occupancy estimates may in fact be irrelevant when far from the involved actors, a single cell with unreliable occupancy estimate may be of utmost importance when close to one of them. In our view, both the probabilistic occupancy grid notion and its calculation as described above capture this concept of locality in perception reliability in a fully transparent and explainable way.

\section{Evaluation and Results}\label{sec:evaluation}
\vspace{-1mm}

The proposed Bayesian CP algorithm was evaluated in a hybrid testing setup combining CARLA simulation with vehicle-in-the-loop experimentation. The objective was to quantify improvements in situational awareness and robustness under varying localization noise levels.

\subsection{Experimental Setup}
\vspace{-1mm}
For each agent, three configurations were evaluated: (1) ego-only perception (no CP), (2) CP with three agents (ego + two others), and (3) CP with all six agents. Metrics were computed per frame and aggregated via area-under-curve (AUC) integration over time, characterizing each agent’s perception consistency across its sequence of presence. The following metrics were analyzed: (1) Field-of-View (FoV) coverage, (2) recall, and (3) precision for occupied and unoccupied cells. All experiments were repeated for localization noise standard deviations of 0, 0.5, 1, 1.5, and 2 m, totaling 90 evaluations per scenario.

\begin{table*}[t]
\centering
\caption{Summary of AUC metrics across all collective perception configurations (averaged over all noise levels).}
\label{tab:evaluation}
\renewcommand{\arraystretch}{1.1}
\setlength{\tabcolsep}{6pt}
\begin{tabular}{l|ccc|ccc|ccc|ccc|ccc}
\hline
\multirow{2}{*}{\textbf{Configuration}} 
& \multicolumn{3}{c|}{\textbf{FoV}} 
& \multicolumn{3}{c|}{\textbf{Occ. Recall}} 
& \multicolumn{3}{c|}{\textbf{Occ. Precision}} 
& \multicolumn{3}{c|}{\textbf{Unocc. Recall}} 
& \multicolumn{3}{c}{\textbf{Unocc. Precision}} \\ 
\cline{2-16}
 & 0\,m & 1\,m & 2\,m 
 & 0\,m & 1\,m & 2\,m 
 & 0\,m & 1\,m & 2\,m 
 & 0\,m & 1\,m & 2\,m 
 & 0\,m & 1\,m & 2\,m \\ 
\hline
\textbf{Ego only (1)} 
& 0.236 & 0.236 & 0.235 
& 0.817 & 0.782 & 0.723 
& 0.939 & 0.814 & 0.742 
& 0.996 & 0.982 & 0.980 
& 0.989 & 0.986 & 0.984 \\

\textbf{CP (3 agents)} 
& 0.643 & 0.643 & 0.642 
& 0.917 & 0.859 & 0.802 
& 0.831 & 0.754 & 0.678 
& 0.988 & 0.983 & 0.978 
& 0.995 & 0.993 & 0.988 \\

\textbf{CP (6 agents)} 
& 0.856 & 0.856 & 0.855 
& 0.938 & 0.874 & 0.808 
& 0.875 & 0.785 & 0.702 
& 0.992 & 0.987 & 0.981 
& 0.996 & 0.993 & 0.990 \\
\hline
\end{tabular}
\vspace{-3mm}
\end{table*}

\subsection{Results and Analysis}

The results, shown in Table ~\ref{tab:evaluation}, summarize the performance of the three configurations under increasing localization noise.

\paragraph{Ego-only perception (no CP)}
In the absence of collective perception, the joint FoV coverage remains limited (approximately 0.23 across all noise levels), reflecting the inherently narrow awareness of single-vehicle sensing. Occupied cell recall and precision degrade as localization noise increases, falling from 0.82 to 0.72 and from 0.94 to 0.74, respectively, between 0 and 2\,m noise. Unoccupied cell metrics remain consistently high ($>0.98$) since free-space estimation is less affected by small localization drifts. This configuration serves as a baseline, emphasizing the vulnerability of ego-only perception to spatial uncertainty.

\paragraph{Collective perception with 3 agents}
When the ego vehicle fuses data with two neighboring connected vehicles, the collective FoV coverage increases almost threefold, averaging around 0.64. Occupied cell recall improves significantly (up to 0.91 at zero noise), remaining above 0.80 even with 2\,m localization noise. Although precision decreases modestly (from 0.83 to 0.68), this is expected due to overlapping FoVs and fusion uncertainty. Unoccupied cell metrics remain stable ($>0.97$ recall and $>0.99$ precision), which is to be expected, since unoccupied cells are of high percentage at all moments. These results confirm that collective Bayesian fusion mitigates the effects of localization error by exploiting redundant observations from neighboring agents. A visualization of the probabilistic occupancy grid with 3 connected agents is shown in the bottom right illustration in Figure \ref{fig:cpmodule}. Darker (lighter) areas correspond to higher (lower) occupancy probabilities. Gray areas correspond to occ. probabilites closer to 0.5, i.e to more uncertain occupancy state.

\paragraph{Collective perception with all agents}
When all six connected vehicles contribute to the collective perception, coverage and reliability reach their highest values. The joint FoV coverage rises to approximately 0.86, approaching full visibility of the area of interest. Occupied cell recall and precision remain high (0.94 and 0.87 at zero noise) and degrade gracefully under noise, maintaining 0.80 recall and 0.70 precision at 2\,m localization uncertainty. Both unoccupied cell recall and precision remain above 0.98 across all noise levels, which is again to be expected due to the reason mentioned in (b). This configuration showcases the scalability and resilience of the proposed Bayesian CP fusion when multiple agents contribute complementary visual and positional data. A visualization of the probabilistic occupancy grid with all the agents connected is shown in the top right illustration in Figure \ref{fig:cpmodule}.

\subsection{Comparative Discussion}
\vspace{-1mm}
Across all noise conditions, the inclusion of collective perception yields a consistent increase in situational awareness and robustness:
\begin{itemize}
    \item \textbf{Spatial coverage:} The joint FoV coverage increases by approximately 260\% from ego-only to full CP configuration.
    \item \textbf{Detection completeness:} Occupied cell recall improves from 0.82 (ego-only) to 0.94 (all agents) at zero noise.
    \item \textbf{Noise tolerance:} At 2\,m noise, performance degradation is significantly lower for CP configurations; recall remains 0.80--0.81 compared to 0.72 in the baseline.
\end{itemize}

While slight decreases in precision occur with an increasing number of fused agents—primarily due to marginally overlapping or redundant detections—the benefits in awareness and robustness substantially outweigh these trade-offs. The results indicate that even small-scale cooperation (three vehicles) provides major performance gains, while full cooperation yields diminishing but still significant improvements.

\section{Conclusion}\label{sec:conclusion}

This paper presented a probabilistic framework and hybrid validation approach for V2X-based Collective Perception (CP) in complex traffic scenarios. The proposed Bayesian fusion algorithm integrates heterogeneous sensor data from multiple connected and autonomous vehicles into a shared occupancy grid, extending perception beyond the ego vehicle’s field of view and quantifying uncertainty at cell level.

A hybrid validation methodology combining CARLA simulation and vehicle-in-the-loop testing enabled reproducible evaluation across virtual and physical domains. Results in a roundabout scenario demonstrated significant gains in spatial coverage and perception robustness: collective FoV coverage increased more than threefold compared to ego-only perception, and occupied-cell recall improved from 0.82 to 0.94 under nominal localization noise.

The framework also supports online reliability assessment through covariance tracking and probabilistic consistency checks. Future work will extend the model to dynamic Bayesian fusion considering communication latency and trust weighting, toward certifiable and explainable deployment of cooperative autonomous systems.

\bibliographystyle{IEEEtran}
\bibliography{biblio}

\end{document}